%% file: neurips_2026_arxiv.tex
\documentclass{article}

\usepackage[preprint, nonatbib]{neurips_2026}

\usepackage[utf8]{inputenc} 
\usepackage[T1]{fontenc}    
\usepackage{hyperref}       
\usepackage{url}            
\usepackage{booktabs}       
\usepackage{amsmath}
\usepackage{amsfonts}       
\usepackage{nicefrac}       
\usepackage{microtype}      
\usepackage{xcolor}         
\usepackage[numbers]{natbib}
\usepackage{graphicx} 
\usepackage{comment}
\usepackage{hyperref}
\usepackage[capitalise,nameinlink]{cleveref}
\usepackage{placeins} 

\title{Track the Noise, Move the World: \\3D-Grounded Motion-Consistent Noise for \\Controllable Video Generation}

\author{%
  \textbf{Long Vu}\thanks{Equal contribution.}\quad
  \textbf{Tan Ngo}\footnotemark[1]\quad
  \textbf{Animesh Karnewar}\quad
  \textbf{Amir Habibian} \\[3pt]
  \textbf{Binh-Son Hua}\thanks{Binh-Son Hua is affiliated with Trinity College Dublin, Ireland. Work done under consultancy capacity.}\quad
  \textbf{Hung Bui}\quad
  \textbf{Minh Hoai Nguyen}\quad
  \textbf{Phong Nguyen-Ha} \\[6pt]
  Qualcomm AI Research\thanks{Qualcomm AI Research is an initiative of Qualcomm Technologies, Inc.} \\[3pt]
  \texttt{\{vlong, tanngo, karnewar, ahabibia,} \\
  \texttt{hson, hungbui, minhhoai, phongnh\}@qti.qualcomm.com}
}

\begin{document}

\input{definitions}
\maketitle

\begin{abstract}
Modern image-and-text-to-video diffusion models can synthesize highly realistic videos by iteratively denoising an initial Gaussian noise tensor conditioned on reference image and text inputs. However, existing approaches still lack precise and unified controllability over both object motion and camera motion within a single generation process. We present \textbf{UniCaMo}, a unified framework that enables simultaneous control of object trajectories and camera viewpoints by directly constructing the input noise of the diffusion model. Specifically, UniCaMo builds a shared \emph{3D-grounded motion-consistent noise space} across latent video frames. Sparse 3D point tracks are used to warp the Gaussian noise of the reference frame along desired object trajectories, while a virtual spherical noise representation provides globally consistent noise values for newly revealed scene regions under camera motion. By combining local track-guided noise warping with global sphere-based noise sampling, UniCaMo maintains geometric and temporal consistency under both object movement and viewpoint changes. Because UniCaMo modifies only the input noise, it requires no auxiliary adapters, control branches, or architectural changes to the underlying video diffusion model. With lightweight LoRA fine-tuning on large pretrained video diffusion models, including Wan 2.1 (14B), UniCaMo achieves state-of-the-art results in both video quality and motion controllability on standard controllable video generation benchmarks.
\end{abstract}

\section{Introduction}

\begin{figure}[t]
\centering
\includegraphics[width=0.9\linewidth]{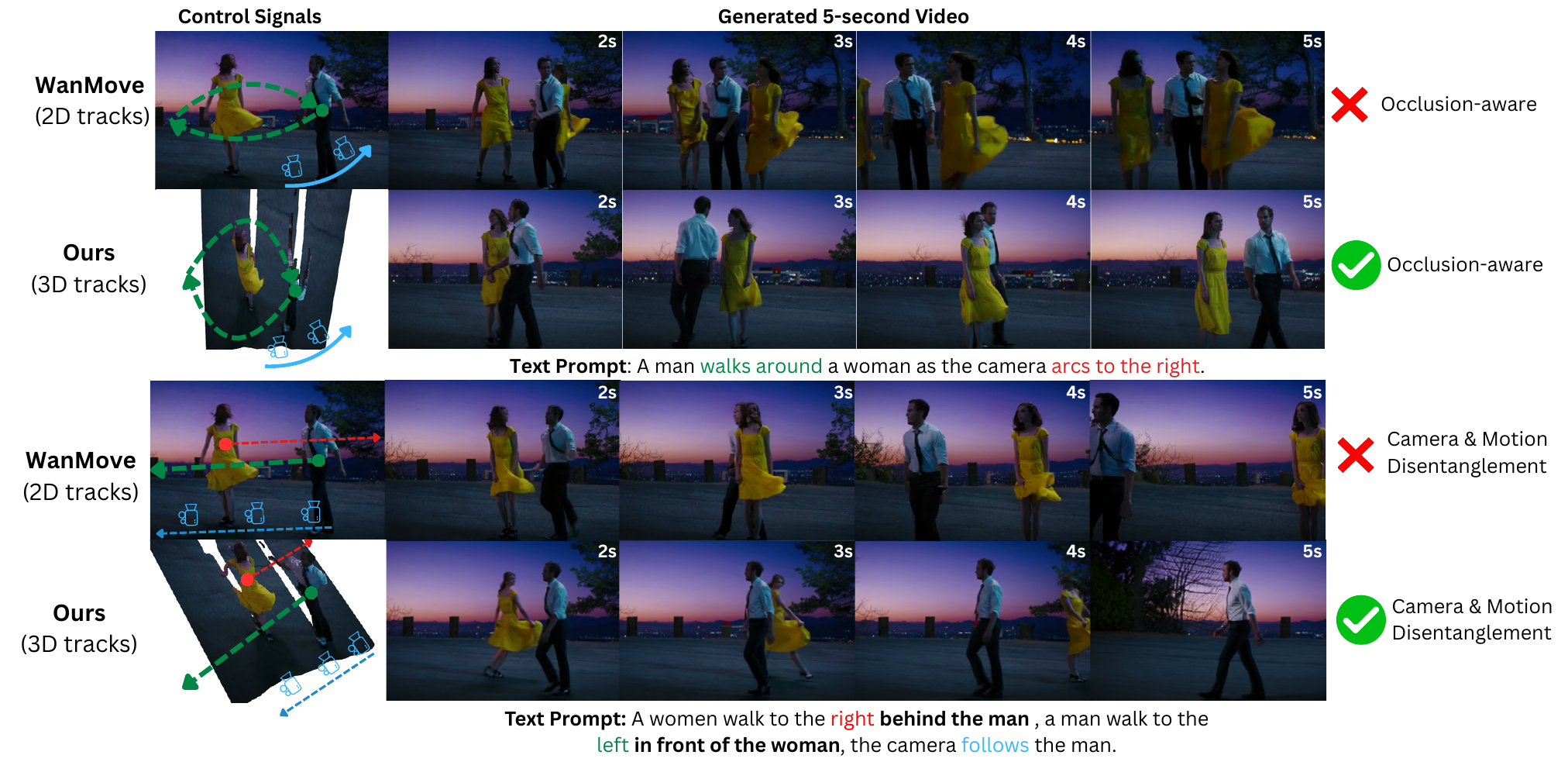}
    
    \caption{UniCaMo jointly controls object + camera motion with 3D tracks and spherical noise, producing coherent, occlusion-aware videos. WanMove~\cite{chu2025wan} relies on 2D guidance, often causing motion/camera ambiguity and implausible occlusions.}
    \label{fig:teaser2}
\end{figure}

Video is a highly expressive visual medium, and precise control over generated content enables significant practical value. A scene is defined not only by its objects, but by their motion and the camera’s movement—e.g., a car on a street looks entirely different when tracked, panned, or viewed from above. Beyond storytelling, controllable video generation is crucial for robotics and embodied AI~\cite{wang2026robovip,jang2025dreamgen}, enabling controllable data synthesis during training and predictive imagination at inference time. Motivated by this, we ask: \emph{can we build a video generation framework that simultaneously and precisely controls object and camera motion without external adapters?}

Controllable video generation has gained increasing attention alongside advances in large-scale video diffusion models~\cite{wanvideo,lightx2v,cogvideo,cogvideox}. Recent works achieve strong control via point tracks, optical flow, camera trajectories, or auxiliary networks~\cite{chu2025wan,he2024cameractrl,bai2025recammaster,burgert2025go,chang2025warped,li2025magicmotion}. However, most methods treat object and camera motion separately—either animating objects from a fixed viewpoint or controlling camera motion with unconstrained scene dynamics. When both are specified, this separation often leads to motion ambiguity and entanglement artifacts. Since object and camera motion are inherently coupled, independent control limits expressiveness. In this work, we explicitly tackle joint object-and-camera motion control, enabling a key capability for modern image- and text-to-video frameworks.

Having established the need for joint motion control, we examine how existing methods specify motion. Prior work~\cite{chu2025wan,li2025magicmotion,wang2024motionctrl} typically relies on sparse 2D point tracks, allowing users to drag image-plane points to indicate object or camera motion. While intuitive, this approach has two key limitations. First, 2D trajectories provide an incomplete and ambiguous description of underlying 3D dynamics: the same image-space motion can arise from multiple combinations of object motion and camera movement, and many motions—such as one person moving around another (Fig.~\ref{fig:teaser2})—are difficult to express without explicit depth or viewpoint information. Second, motion constraints are usually injected late in the generation pipeline, when visual content has already emerged. This leaves limited flexibility for resolving motion ambiguity in a geometrically coherent way while maintaining realism and temporal consistency. Together, these issues often lead to imprecise, entangled, or geometrically inconsistent motion control.

This observation motivates a different perspective: instead of injecting ambiguous motion guidance near the output, can we impose motion constraints directly at the input? Modern image-and-text-to-video diffusion models generate videos by iteratively denoising an initial Gaussian noise tensor conditioned on a reference image~\cite{wanvideo}. This suggests a reformulation of controllable video generation as the problem of constructing an input noise tensor that already encodes the desired scene dynamics. Rather than controlling the generation process itself, we control its input. Our approach is thus related to recent noise-warping methods~\cite{chang2025warped,burgert2025go}, which achieve strong object motion control by warping object-associated noise across frames. However, extending these methods to joint object-and-camera motion is fundamentally challenging. Under a static camera, object motion creates small, localized voids that diffusion models can often infer and inpaint using surrounding context. In contrast, camera motion exposes large unseen regions with no corresponding noise from the reference view. As the viewpoint diverges, these missing regions grow, making noise warping from a single reference-frame fundamentally insufficient for preserving temporal consistency and geometric coherence.

To address these challenges, we propose a motion-consistent noise representation that remains geometrically coherent under both object and camera motion. Instead of defining noise only on the reference image plane, we lift it into a shared spherical 3D representation parameterized by Gaussian noise, providing consistent noise values for all viewing directions. As the camera moves, each frame samples from this shared sphere based on its viewpoint, enabling temporally coherent synthesis of newly revealed regions. Within this space, user-specified point tracks locally warp noise along object trajectories, ensuring consistent object motion under changing viewpoints. This 3D formulation allows users to specify both object and camera motion directly in 3D, avoiding image-space ambiguity. Combining sphere-based noise sampling with track-guided warping, our method—UniCaMo—constructs a unified 3D-grounded motion-consistent noise space that enables precise joint control of object and camera motion while modifying only the input noise. UniCaMo requires no architectural changes or auxiliary adapters, supports lightweight LoRA fine-tuning of pretrained models such as Wan 2.1 (14B)~\cite{wanvideo}, adds only 0.2s per sample, and achieves state-of-the-art video quality and controllability~\cite{chu2025wan}.

In summary, our main contributions are: \textbf{(1)} we propose {UniCaMo}, a novel motion-consistent noise representation that combines 3D point tracks with sphere-projected Gaussian noise to jointly encode object and camera motion in a unified 3D space; \textbf{(2)} UniCaMo requires no auxiliary adapters or trainable control modules, introducing only negligible warping overhead (approximately 0.2s per sample) while preserving the runtime efficiency of the base diffusion architecture; and \textbf{(3)} extensive experiments on standard controllable video generation benchmarks, including MoveBench~\cite{chu2025wan}, demonstrate state-of-the-art controllability and high visual quality, particularly in scenarios involving complex coordinated object and camera motion.

\section{Related Work}
\label{sec:related}

Diffusion-based video generation has advanced from early 3D U-Nets~\cite{lumiere,svd,videocrafter1,animatediff,vdm2022,imagen,moviegen} to large-scale latent diffusion models with stronger spatiotemporal modeling. Transformer-based foundation video models further improve long-range coherence and fidelity via spatiotemporal attention~\cite{jin2024pyramidal,cosmosNVIDIA,cogvideo,hunyuanvideo,wanvideo,cogvideox,cogvideox,hunyuanvideo}. We build on efficient pretrained models such as Wan~\cite{wanvideo}, and achieve controllable generation through structured latent-noise modeling and lightweight fine-tuning—without architectural changes.


\noindent\textbf{Motion and camera control for video generation.} 
Motion-controlled I2V methods differ by motion cues and how they are injected: masks/boxes/keypoints enables coarse motion cues, while flow or trajectories provide finer guidance~\cite{jain2024peekaboo,wang2024boximator,li2025magicmotion,zhong2024posecrafter,burgert2025go,koroglu2025onlyflow,geng2025motion,yin2023dragnuwa,zhang2025tora}. Among these methods, many rely on ControlNet-style adapters with higher cost, whereas Wan-Move~\cite{chu2025wan} edits conditioning features without architecture changes. 
%
%
Camera control in video diffusion is typically achieved by conditioning on 6DoF poses/trajectories, injected via dedicated modules~\cite{he2024cameractrl,wang2024motionctrl,bahmani2025ac3d}. Related work enforces cross-view consistency (e.g., CVD) or performs camera-only V2V “reshoots,” but these methods generally preserve object dynamics and do not support simultaneous object-motion editing~\cite{kuang2024collaborative,bai2025recammaster,ren2025gen3c,bahmani2025lyra}.


\noindent\textbf{Warped noise for controllable video generation.}
A promising direction is to control diffusion by using structured initialization noise. Noise-warping methods create temporally correlated noise by warping Gaussian noise with motion fields (e.g., optical flow), injecting temporal structure while preserving per-frame spatial Gaussianity. Warped-noise priors such as HIWYN~\cite{chang2025warped} highlight the importance of this constraint, though large deformations can introduce artifacts and overhead. Go-with-the-Flow~\cite{burgert2025go} makes warped noise efficient and shows motion control can be added by changing only the noise/data pipeline, treating the diffusion model as a black box. Recent analysis further finds that training with warped noise promotes useful equivariances and can enable few-step sampling~\cite{liu2025equivdm}.

\noindent\textbf{Training-free methods.}
Training-free methods control video diffusion at inference by editing noise/latents, attention, or denoising schedules~\cite{camtrol,dvs,singer2025time}. Examples include CamTrol and DVS for camera/viewpoint control, and Time-to-Move for motion via region-wise schedules, but such methods often struggle with long-horizon, multi-object, or precise control~\cite{camtrol,dvs,singer2025time}.

\noindent\textbf{Unified control.}
Existing methods with joint camera and object control often relies on 3D adapters or 3D primitives, increasing runtime cost~\cite{wang2024boximator,li2025magicmotion,geng2025motion,shi2024motion,chu2025wan,editbytrack,zheng2026versecrafter}. 
Particularly, trajectory-based methods aim for unified control by using background tracks for camera motion and foreground tracks for object motion. Edit-by-Track leverages 3D point tracks to resolve depth/occlusion in joint edits, while other systems use geometry-aware adapters or richer 3D states, typically adding conditioning modules and runtime~\cite{editbytrack,zheng2026versecrafter}. In contrast, we target joint camera+object control via structured noise: we add no inference-time adapters~\cite{wang2024imageconductor,wang2024motionctrl} and avoid dense flow warping~\cite{burgert2025go} by combining sparse 3D tracks with sphere-projected noise to handle disocclusions while preserving per-frame Gaussianity.


\section{UniCamMo: a Method for Joint Camera and Object Motion Control}
\label{sec:method}
UniCaMo is built upon an image-and-text-to-video framework such as Wan~\cite{wanvideo}. 
It accepts the following input: (i) a source image $I$ and text prompt $y$, (ii) a sparse set of target 3D point tracks per video frame $\mathcal{T}=\{ \mathbf{x}_i^t \in \mathbb{R}^3 \}_{i=\overline{1,N},\,t=\overline{1,T}}$, and (iii) target per-frame camera poses $\mathcal{C}=\{\mathbf{P}^t\}_{t=\overline{1,T}}$, with $N$ being the number of point tracks per frame and $T$ the number of video frames. Note that during inference, both $\mathcal{T}$ and $\mathcal{C}$ are defined by the user using a graphical interface.


During training, we construct the input based on original video data as follows. For each video, we extract the first frame and use it as the source image $I$, and apply video captioning to extract text prompt $y$. We then apply point tracks, camera pose and depth estimation on the video to obtain point tracks $\mathcal{T}$ and pose $\mathcal{C}$. During inference, we assume that the target 3D point tracks $\mathcal{T}$ and target camera poses $\mathcal{C}$ are specified by the user through a graphical user interface.
Our goal is to generate a video with $T$ frames initialized from the content of the source image $I$ with motion dynamics following the target object and camera motion encoded by point tracks $\mathcal{T}$ and camera pose~$\mathcal{C}$.


\begin{figure}[t]
    \centering
  \includegraphics[width=0.8\textwidth]{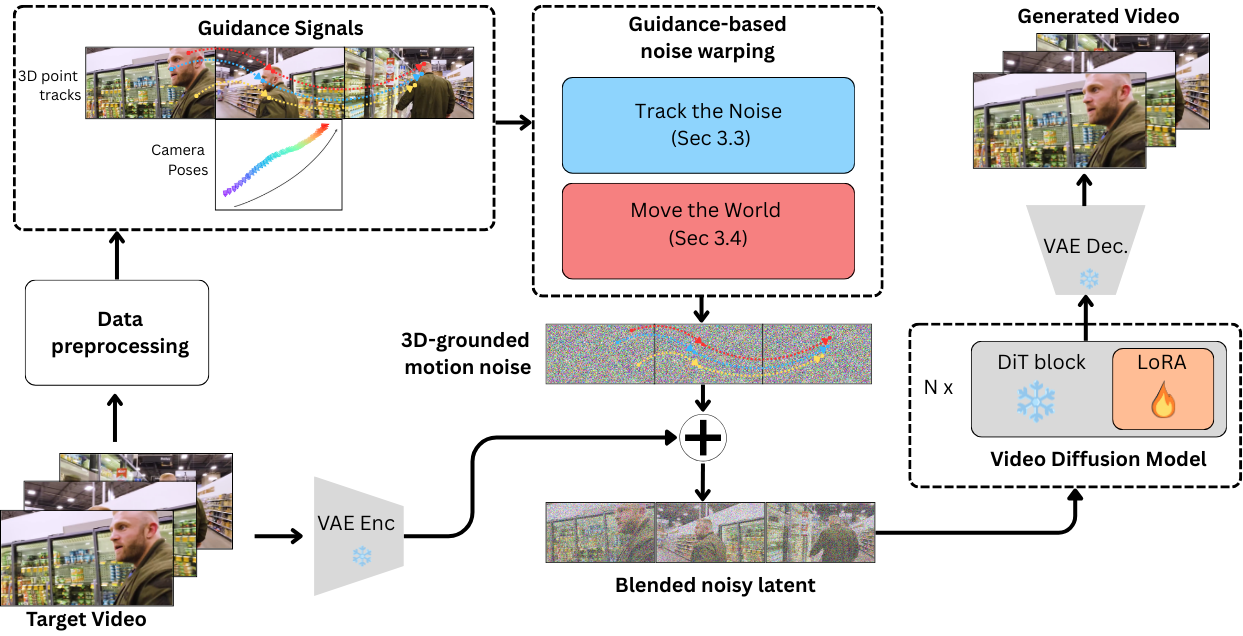}
  \caption{Our framework trains a video diffusion model by warping initialization noise using 3D point tracks and camera poses (via Track-the-Noise and Move-the-World) to create motion-consistent 3D-grounded latents, enabling independent or joint control of object and camera motion at inference.
  }
  \label{fig:overall_pipeline}
\end{figure}

Fig.~\ref{fig:overall_pipeline} summarizes our training pipeline. We first perform data pre-processing to obtain depth maps, camera poses, and 3D point tracks of the target video using off-the-shelf 3D models such as ViPE~\cite{huang2025vipe} or TAPIP3D~\cite{tapip3d}. We then apply a noise warping technique to obtain our 3D-grounded motion-consistent noise through two steps: (1) Track-the-Noise (Sec.~\ref{sec:tracknoise}), which creates partial motion-consistent noise that captures the motion inferred from 3D point tracks, and (2) Move-the-World (Sec.~\ref{sec:sphere}), which samples and propagates Gaussian noise on a virtual 3D sphere to produce view-consistent noise along a target camera trajectory. The combined representation results in 3D-grounded motion-consistent noise that can be used to initialize the latents of a video diffusion model~\cite{wanvideo} to generate the final output video.


\subsection{Track the Noise: Noise Initialization using 3D Point Tracks}
\label{sec:tracknoise}

\begin{figure}[t]
    \centering
  \includegraphics[width=0.8\textwidth]{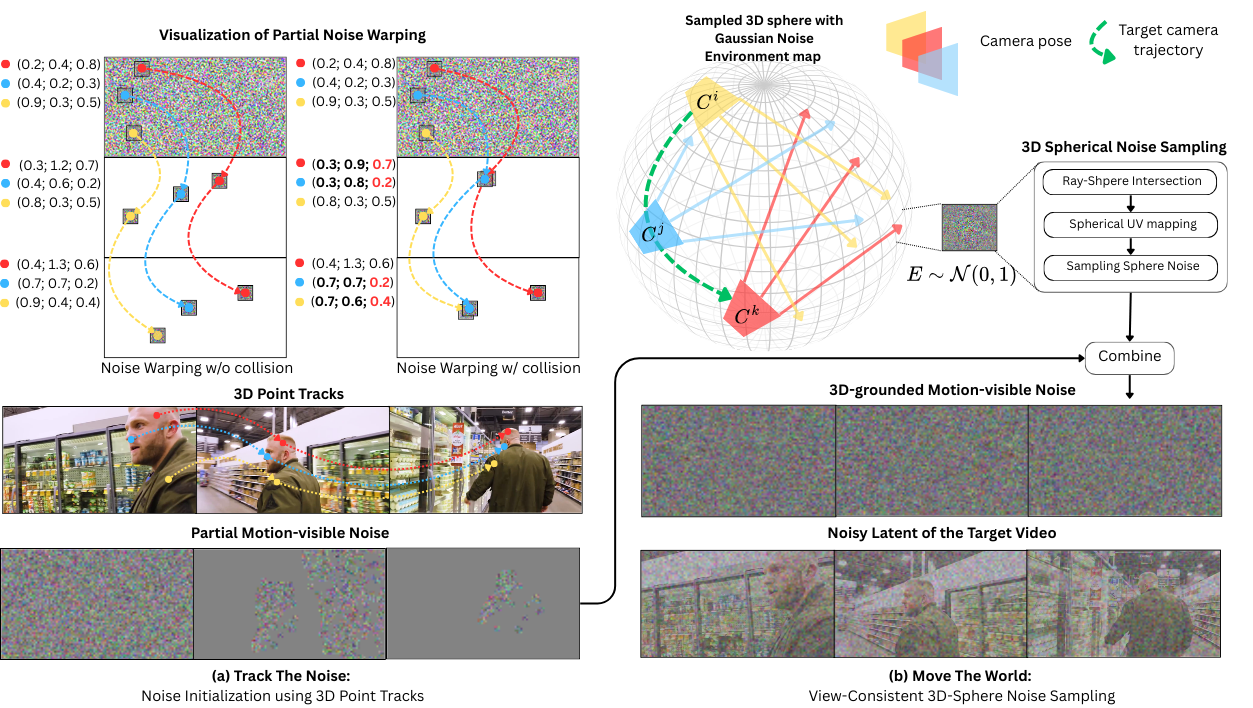}
  \caption{Noise representation in our two key steps, Track the Noise (a) and Move the World (b). (a) We track 3D points (see example triplets of X, Y, Z values) and resolve potential collisions in their trajectories using a depth test, favoring contributions from points closer to the camera. This step results in initial partial motion-visible noise. (b) We create a virtual 3D sphere with Gaussian‑textured noise and sample view‑consistent noise according to the target camera poses. By combining the outputs of (a) and (b), we obtain 3D-grounded, motion-consistent noise that replaces the randomly sampled noise used in standard video diffusion models.}
  \label{fig:noiselatents}
\end{figure}

\noindent\textbf{Motivation.} Conventional video diffusion models~\cite{wanvideo,cogvideox} initialize generation with a single Gaussian noise latent, which effectively behaves as frame‑wise uncorrelated noise, lacking any temporal structure.
Yet real videos exhibit strong motion correlations, recoverable via dense optical flow~\cite{burgert2025go} or sparse point tracks~\cite{editbytrack}. Leveraging this as an inductive bias, we partially warp the initial noise using the estimated motion (Fig.~\ref{fig:noiselatents}(a)), producing a motion‑consistent noise prior that guides the model toward coherent video dynamics.

\noindent\textbf{Partial Noise Warping.} Let $\mathcal{T} = \{\mathbf{x}_i^t \in \mathbb{R}^3 \}$ denote the set of world-space 3D point tracks across video frames.
Compared to dense optical flow \cite{raft}, sparse 3D tracks provide explicit depth cues and are robust to occlusion and viewpoint changes. 
We map the point tracks to latent feature map locations by first projecting them onto the image plane using the target camera poses $\{\mathbf{P}^t\}$, and then rescaling the projected coordinates to the latent feature map resolution. We use the scale ratio $W_{lat}/W$ and $H_{lat}/H$, where $W_{lat}, H_{lat}, W, H$ are latent and image dimensions, respectively (for Wan 2.1 14B model~\cite{wanvideo}, this scale ratio is 4). Through 3D-to-2D projection, we obtain the 2D pixel trajectories be $\{ \mathbf{u}_i^t \in \mathbb{R}^2 \}$, corresponding to the 3D point tracks.

By initializing the latent representation of the first frame as i.i.d. Gaussian noise, we now propagate the noise to subsequent frames following the pixel trajectories $\{ \mathbf{u}_i^t \}$ constructed from 3D point tracks. 
We simply transfer (i.e., copy and write) noise within a square patch with dimension $2R - 1$ centered at each pixel in the trajectories from the first frame to the target frames. 
By default, we set $R = 2$, resulting in patch of size 3x3.
As the 3D point tracks encode motion in the video, this results in a video latent representation that exhibits motion-consistent noise, forming a strong guidance for generating motion dynamics during video synthesis. 

\myheading{Depth-aware Collision Tracking}. 
When multiple point trajectories overlap and project to the same latent pixel, naive composition leads to ambiguous assignments and may introduce inconsistent motion cues. 
We therefore resolve collisions using a depth-aware z-buffer rule induced by the target tracks $\mathcal{T}$ and camera poses $\mathcal{C}$. 
Among all candidates mapped to a latent pixel, we select the contribution with the smallest camera depth (i.e., the point closest to the camera) and discard the rest. 
This choice enforces a physically plausible occlusion ordering and leverages the explicit depth information provided by 3D tracks to disambiguate visibility during noise propagation.

In contrast, prior motion-control pipelines often adopt simple but less effective heuristics for handling overlaps, such as selecting one contributor without geometric reasoning (e.g., by arbitrary ordering) \cite{chu2025wan}, or aggregating multiple contributors through averaging (sometimes followed by re-normalization to restore Gaussian statistics) \cite{burgert2025go}. 
By explicitly enforcing depth-consistent compositing, our construction yields geometrically coherent structured noise signals, improving the stability of the noise representation under occlusions and viewpoint changes.



\subsection{Move-the-World: View-Consistent 3D-Sphere Noise Sampling}
\label{sec:sphere}


\noindent\textbf{Motivation.} 
As the Track‑the‑Noise step transports first‑frame noise along point‑track trajectories, some regions remain unconstrained—especially areas that become newly visible as camera motion reveals previously unseen parts of the scene.
Filling these gaps with random Gaussian noise provides no motion cues, while copying first‑frame noise introduces incorrect parallax.
To address this, we introduce Move‑the‑World, which represents the scene as a 3D bounding sphere coated with Gaussian noise. The camera moves inside this sphere, and projecting the sphere’s noise into each view produces view‑consistent latents that preserve spatial Gaussianity (Fig.~\ref{fig:noiselatents}(b)). 

\noindent\textbf{3D Sphere Parameterization.}
Constructing a bounded 3D sphere requires defenition on the center position and radius.  
In particular, we set the sphere center to the camera center of the first frame and choose a radius $r$ such that the sphere encloses the entire camera trajectory across frames while satisfying a conservative depth bound $d_{\max}$. The radius $r$ is define as follows: $
r = \max\!\left( \tau \cdot \max_{t}\left\|\mathbf{C}^t-\mathbf{C}^1\right\|_2,\; d^{0}_{\max}\right)$ where $\mathbf{C}^t \in \mathbb{R}^3$ denote the camera center at frame $t$, $d^{0}_{\max}$ is the furthest depth value of the input image and $\tau = 1.2$. 
We represent the sphere surface by a latitude-longtitude parameterization, which can be captured by a randomly sampled Gaussian noisy texture map  $\mathbf{E} \in \mathbb{R}^{C \times H_{\text{env}} \times W_{\text{env}}}$, where $H_{\text{env}}$ and $W_{\text{env}}$ are proportional to the video resolution scaled by a factor $k$ ($k = 4$), and $C$ matches the latent channel count of the video diffusion model~\cite{wanvideo,cogvideox}.

\noindent\textbf{Ray-sphere Intersection \& Noise Sampling. } To obtain 3D‑consistent noisy latents for each camera view, we cast rays from the camera through its view frustum and compute their intersections with a bounding sphere using a closed‑form ray–sphere intersection formula. For every ray that hits the sphere surface, we determine the latitude and longitude of the intersection point. These spherical coordinates are then used to perform nearest‑neighbor grid sampling on the previously defined noisy texture map $E$. This procedure produces noise latents that remain spatially Gaussian within each frame while staying consistent across different viewpoints. Additional details on the ray–sphere intersection computation can be found in the appendix section. 

\noindent\textbf{Noise Composition.} To combine the motion‑consistent noises produced independently by Track‑the‑Noise and Move‑the‑World, we apply a composition step that aggregates their outputs into a single final noise latent. We begin by initializing the final latent with the partial motion‑consistent noise obtained from the Track‑the‑Noise stage. Then, for regions where this step provides no coverage, we fill in the missing areas by copying the corresponding values from the Move‑the‑World output. This strategy yields a complete latent that is both 3D‑grounded and motion‑consistent, making it well‑suited for the subsequent video generation with the diffusion model. 

 


\subsection{Training and Inference}
\label{sec:training_inference}
\noindent\textbf{Training data.}
We curate 400K training clips from DynPose100K++, WSDG-1M, and 4DNeX~\cite{huang2025vipe,chen20254dnex}. For each clip, we precompute and cache offline the signals needed for noise construction: camera intrinsics/extrinsics and depth from ViPE~\cite{huang2025vipe}, and sparse 3D point tracks from TAPIP3D~\cite{tapip3d}. These signals are used only to build the motion-consistent noise initializer (not as additional conditioning inputs). To standardize captions across sources, we re-caption all clips with Qwen2.5-VL~\cite{bai2025qwen3}, following the Wan-Move prompt-extension protocol to improve prompt consistency~\cite{chu2025wan}.

\noindent\textbf{Training details.} We fine-tune a pretrained Wan I2V backbone~\cite{wanvideo} using flow matching to predict the velocity field that transports noise samples to the data distribution. We start from Wan2.1-I2V-A14B weights and apply LoRA only to the DiT denoiser, freezing the VAE and all encoders. We use LoRA rank 64 with $\alpha{=}1$, and train for 10K steps in bfloat16 with gradient checkpointing on 32 NVIDIA H100s (batch size 1/GPU) for ~3 days. To improve robustness to varying guidance density, we randomly sample 512–1024 points from the extracted 3D tracks during training.


\noindent\textbf{Inference.}
Given a single input image, we estimate monocular depth to obtain a 3D point cloud. The user selects target objects via a mask (manual or SAM~\cite{carion2025sam3segmentconcepts}), then specifies 3D point tracks for these objects over time and optionally a camera trajectory (e.g., via a simple 3D GUI). These signals are passed to our Guidance-based Noise Warping module to construct 3D-grounded, motion-consistent noise. We generate videos with classifier-free guidance ($\gamma = 5.0$). UniCaMo adds no extra parameters and incurs only 0.2s warping overhead per sample, preserving the base model’s runtime. Additional qualitative results and demos are provided in the supplement/project page.


%
%

\section{Experiments}
\label{sec:experiments}

\subsection{Experimental Setup}
 \noindent\textbf{Baselines.} 
We evaluate UniCaMo on controllable I2V generation, focusing on joint camera–object motion. We compare against adapter-based controllers (MagicMotion, Tora, ImageConductor, LeviTor), feature-warping Wan-Move, and flow-based noise-warping Go-with-the-Flow~\cite{li2025magicmotion,zhang2025tora,wang2024imageconductor,wang2025levitor,chu2025wan,burgert2025go}. All methods use the same inputs and we report averages over full benchmark splits.
\noindent\textbf{Benchmarks and Metrics.}
We report results on the standard controllable video benchmark, namely MoveBench \cite{chu2025wan}. 
We evaluate along two axes:
\textbf{(i) visual quality} using FID \cite{fid}, FVD \cite{fvd} and reconstruction-style metrics PSNR, SSIM \cite{ssim}, and
\textbf{(ii) control fidelity} using motion accuracy metrics such as end-point error (EPE) for trajectories. MoveBench~\cite{chu2025wan} provides only 2D tracks and first-frame masks, so we recover the required 3D tracks and camera poses using off-the-shelf TAPIP3D and ViPE~\cite{tapip3d,huang2025vipe}. Using the masks, we sample \(M\) object points (proportional to mask size) and \(N\!-\!M\) background points to match our training track count. To provide an apple-to-apple comparison under matched track sources, we additionally evaluate a \textit{Wan-Move + TAPIP3D tracks} variant, in which we project the same 3D tracks used by UniCaMo onto the 2D image plane and feed them as input to Wan-Move.

\subsection{Qualitative and Quantitative Evaluation}
\begin{figure}[t]
    \centering
  \includegraphics[width=0.8\textwidth]{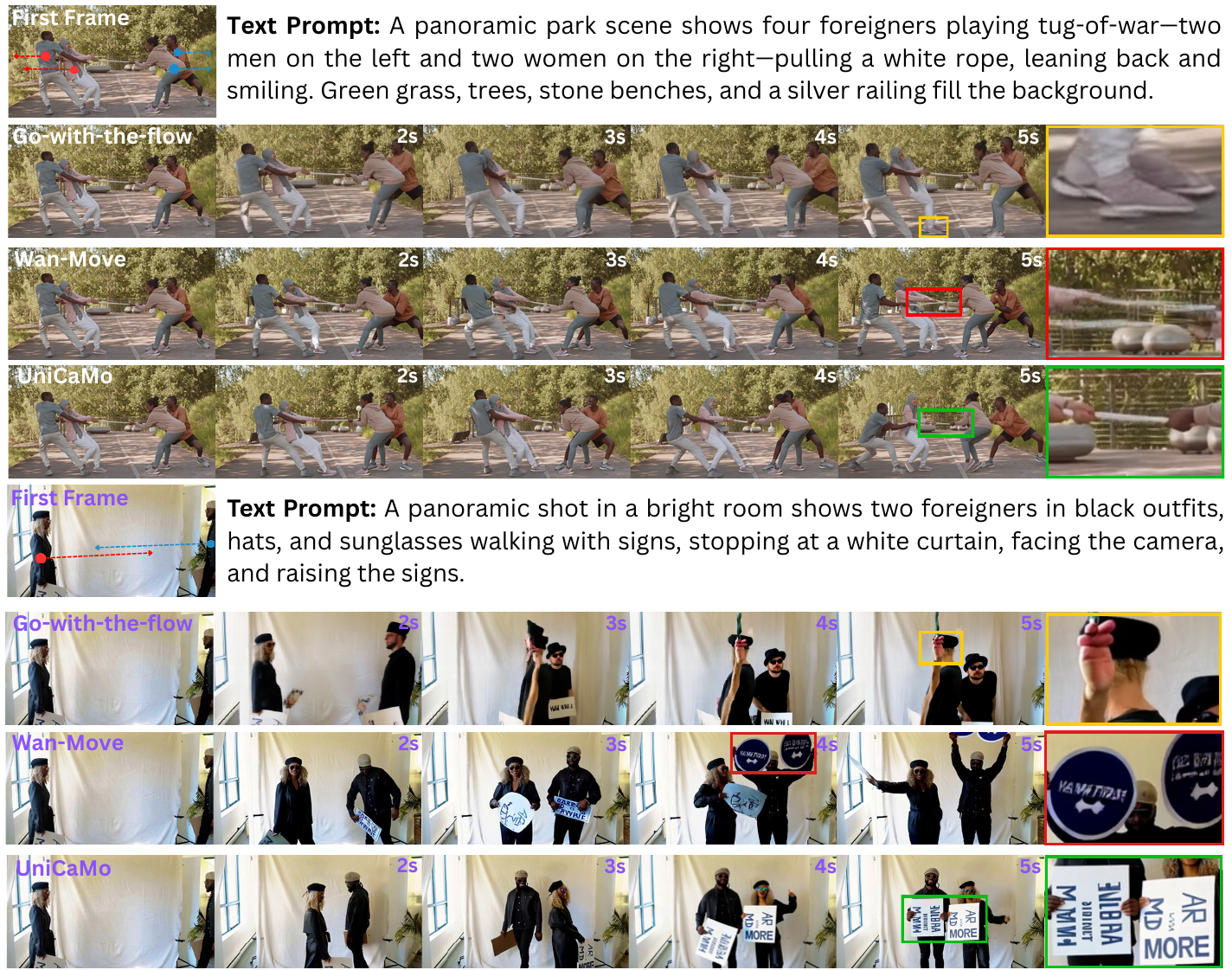}
  \caption{Qualitative results on MoveBench between UniCaMo and recent approaches~\cite{chu2025wan,li2025magicmotion,zhang2025tora}. Our method consistently outperforms other baselines and manages to follow the target motion and camera. Visual artifacts are shown in red boxes.}
  \label{fig:move-bench}
\end{figure}

Figure~\ref{fig:move-bench} shows representative examples in MoveBench under joint camera and object motion control.
Overall, UniCaMo generates temporally coherent videos that better preserve scene structure while following the specified object trajectories and camera motion.
Compared to another baselines, our results exhibit fewer visuals artifacts and manages to follow the conditioning camera and object motions. Go-with-the-flow~\cite{burgert2025go} produces visible artifacts in small detail like human's heah or feet, Wan-Move~\cite{chu2025wan} while exhibits less artifacts in small detail, failed to generate motion that follow user's intention. Please see the included videos in the supplementary material for more scene details. 


\Cref{fig:dynamic-bench} demonstrates the generalization of our method on unseen in-the-wild samples with complex geometry and trajectory. 
In the first example, we expect the white van to move leftward out of the video frame while the camera moves to the right. In this case, Wan-Move~\cite{chu2025wan} failed to generate plausible results. Our 3D-grounded motion-consistent noise representation represents such motion dynamics based on user-provided point tracks and camera trajectories, resulting in high-quality motion content.
In the second example, similarly, Wan-Move can pan the camera to the right but does not generate the cars in the correct moving direction. 
Contrastively, our method can generate realistic flows of the cars in the crowded street. 

\begin{figure}[t]
    \centering
  \includegraphics[width=0.8\textwidth]{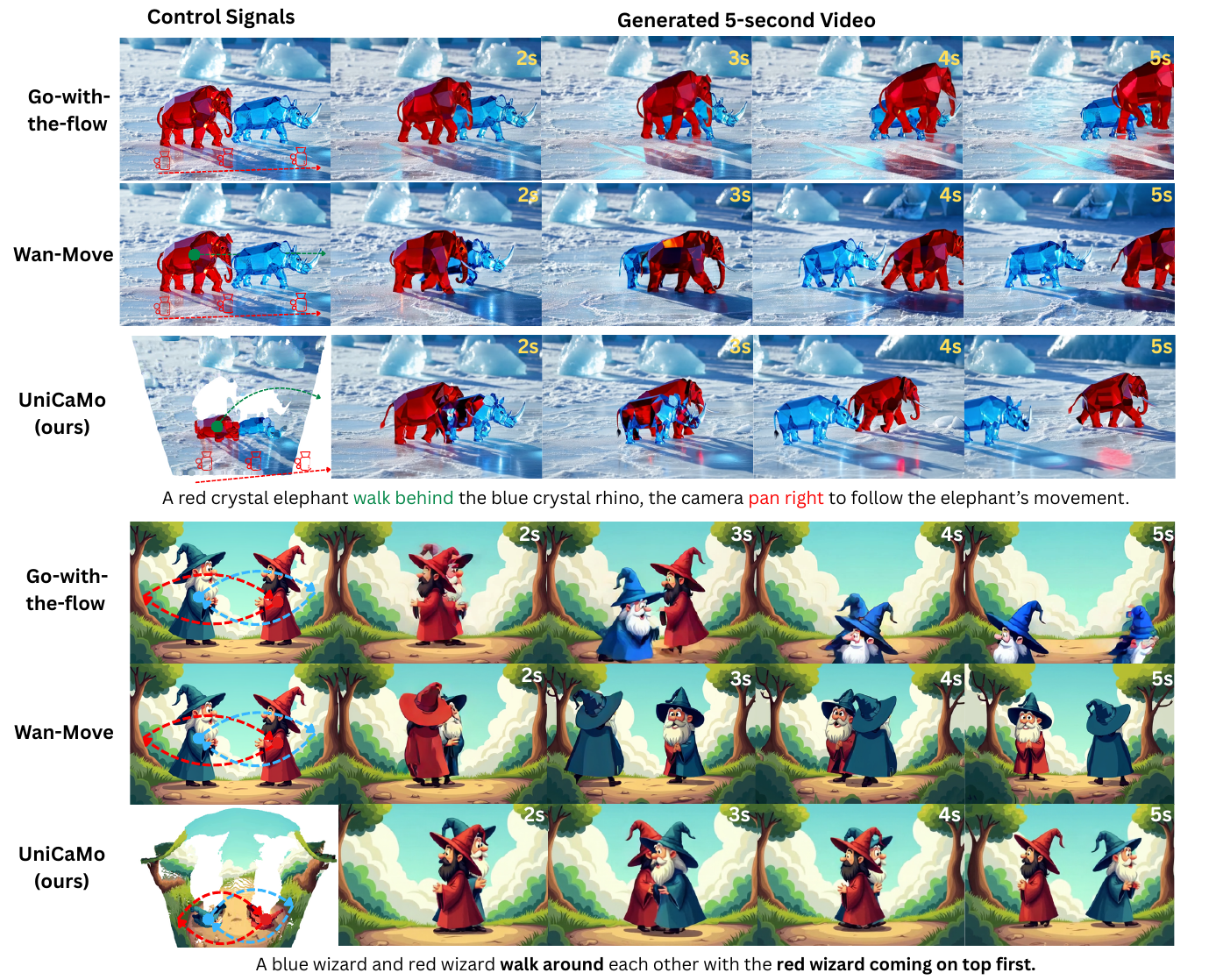}
  \caption{Qualitative results on complex in-the-wild samples between our proposed UniCaMo and Wan-Move~\cite{chu2025wan}, Go-with-the-flow~\cite{burgert2025go}}
  \label{fig:dynamic-bench}
\end{figure}

\begin{figure}[t]
    \centering
  \vspace{-0.3cm}
  \includegraphics[width=0.8\textwidth]{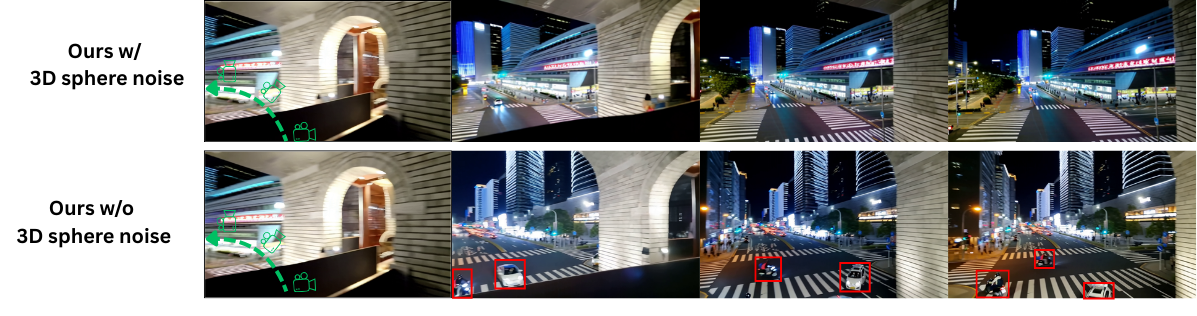}
  \caption{Ablation results on the 3D sphere noise. Without 3D-consistent sphere noise sampling, the model tends to hallucinate random objects (red boxes) with implausible motion trajectories.}
  
  \label{fig:sphere}
\end{figure}


\begin{table}[t]
\centering
\caption{Evaluation results on MoveBench. We report quantitative comparisons on both the single-object and multi-object motion splits. ``--'' indicates the baseline does not provide results for that split. UniCaMo achieves best overall performance across. Best in \textbf{bold}; second-best \underline{underlined}.}
\label{tab:movebench}
\setlength{\tabcolsep}{4pt}
\resizebox{\textwidth}{!}{%
\begin{tabular}{l|ccccc|ccccc}
\toprule
& \multicolumn{5}{c|}{\textit{MoveBench (Single-object motion)}}
& \multicolumn{5}{c}{\textit{MoveBench (Multi-object motion)}} \\
Methods
& FID $\downarrow$ & FVD $\downarrow$ & PSNR $\uparrow$ & SSIM $\uparrow$ & EPE $\downarrow$
& FID $\downarrow$ & FVD $\downarrow$ & PSNR $\uparrow$ & SSIM $\uparrow$ & EPE $\downarrow$ \\
\midrule
ImageConductor~\cite{wang2024imageconductor}
& 34.51 & 424.1 & 13.4 & 0.49 & 15.66
& 77.5  & 764.5 & 13.9 & 0.51 & 9.80 \\
LeviTor~\cite{wang2025levitor}
& 18.12 &  98.8 & 15.6 & 0.54 &  3.40
& --    & --    & --   & --   & --   \\
MagicMotion~\cite{li2025magicmotion}
& 17.53 &  96.7 & 14.9 & 0.56 &  3.20
& --    & --    & --   & --   & --   \\
Go-with-the-Flow~\cite{burgert2025go}
& 12.49 & 216.9 & 15.8 & 0.62 &  3.04
& 35.7  & 399.4 & \underline{16.9} & 0.60 & 3.71 \\
Tora~\cite{zhang2025tora}
& 22.57 & 100.4 & 15.7 & 0.55 &  3.30
& 53.2  & 350.0 & 14.5 & 0.54 & 3.50 \\
Wan-Move~\cite{chu2025wan}
& 12.23 & \underline{83.5} & 17.8 & 0.64 &  2.60
& 28.8  & \underline{226.3} & 16.7 & 0.62 & 2.20 \\
Wan-Move + TAPIP3D tracks~\cite{chu2025wan,tapip3d}
& \underline{12.11} &  87.1 & \underline{18.1} & \underline{0.67} & \underline{2.54}
& \underline{27.6}  & 294.8 & 16.5 & \underline{0.65} & \underline{2.14} \\
UniCaMo (Ours)
& \textbf{10.36} & \textbf{78.6} & \textbf{18.4} & \textbf{0.73} & \textbf{2.30}
& \textbf{26.1}  & \textbf{215.2} & \textbf{18.15} & \textbf{0.72} & \textbf{1.92} \\
\bottomrule
\end{tabular}%
}
\vspace{-0.3cm}
\end{table}
\noindent\textbf{Single-object motion control.} Table~\ref{tab:movebench} (left block) summarizes quantitative results on the single-object split of MoveBench.
UniCaMo achieves the best overall performance across both video quality and control fidelity metrics.
Our model yields substantially improved quality and control compared to prior motion-control methods, with lower FID/FVD and higher PSNR/SSIM while reducing trajectory error (EPE).
In particular, on MoveBench we obtain FID $10.36$ and FVD $78.6$, improving over Wan-Move (FID $12.2$, FVD $83.5$) and trajectory-based baselines in the same evaluation protocol.
We also observe consistent gains in SSIM and EPE (e.g., SSIM $0.73$ and EPE $2.3$ for our model versus SSIM $0.64$ and EPE $2.6$ for Wan-Move).
Notably, Go-with-the-Flow~\cite{burgert2025go} did not produce meaningful videos under MoveBench's joint-control protocol despite repeated attempts, indicating that flow-based warped-noise priors do not transfer to this evaluation.

\noindent\textbf{Multi-object motion control.}
As MoveBench includes 192 cases with annotated multi-object motion, we further evaluate UniCaMo against baselines~\cite{wang2024imageconductor,zhang2025tora,burgert2025go,chu2025wan} on this challenging setting, as presented in Table~\ref{tab:movebench} (right block).
Our method achieves the lowest FVD and reduced EPE compared to other methods, highlighting its precise adherence to motion constraints in more complex scenarios.

\noindent\textbf{Supplementary videos.} Please check our provided HTML page to see the videos and comparisons.



\subsection{Ablation Studies}
\label{sec:ablation}

\begin{table}[t]
\centering
\caption{Ablation studies on UniCaMo on the MoveBench~\cite{chu2025wan} single-object motion evaluation set. We report VBench-I2V~\cite{huang2023vbench} scores together with PSNR, SSIM, and End Point Error (EPE). \textit{(a)} compares our method with and without the sphere-projected noise of Move-the-World; \textit{(b)} sweeps the degradation rate $\theta$ used to blend constructed motion-noise with pure Gaussian noise.}
\label{tab:movebench_ablation}
\renewcommand{\arraystretch}{1.15}
\setlength{\tabcolsep}{4pt}
\resizebox{\textwidth}{!}{%
\begin{tabular}{l|ccc|ccccccc}
\toprule
\textbf{Setting}
& \shortstack{PSNR$\uparrow$}
& \shortstack{SSIM$\uparrow$}
& \shortstack{EPE$\downarrow$}
& \shortstack{Subject\\Consistency$\uparrow$}
& \shortstack{Background\\Consistency$\uparrow$}
& \shortstack{Aesthetic\\Quality$\uparrow$}
& \shortstack{Imaging\\Quality$\uparrow$}
& \shortstack{Overall\\Consistency$\uparrow$}
& \shortstack{Temporal\\Flickering$\uparrow$}
& \shortstack{Motion\\Smoothness$\uparrow$} \\
\midrule
\multicolumn{11}{c}{\textit{(a) Effect of sphere-projected noise (Move-the-World)}} \\
\midrule
Ours w/ sphere
& \textbf{19.41} & \textbf{0.79} & \textbf{2.21}
& \textbf{95.29} & \textbf{96.38} & \textbf{53.41}
& 69.58 & 21.65 & \textbf{98.31} & \textbf{99.12} \\
Ours w/o sphere
& 19.10 & \textbf{0.78} & 2.42
& 94.79 & 96.11 & 52.94
& \textbf{70.05} & \textbf{21.72} & 98.11 & 98.92 \\
\midrule
\multicolumn{11}{c}{\textit{(b) Effect of degradation rate $\theta$}} \\
\midrule
$\theta = 0.0$
& \textbf{19.46} & \textbf{0.79} & \textbf{2.19}
& \textbf{95.40} & 96.35 & 53.39
& 69.51 & \textbf{21.75} & \textbf{98.33} & \textbf{99.13} \\
$\theta = 0.2$
& 19.41 & \textbf{0.79} & 2.21
& 95.29 & \textbf{96.38} & 53.41
& 69.58 & 21.65 & 98.31 & 99.12 \\
$\theta = 0.5$
& 19.02 & 0.78 & 2.49
& 95.08 & 96.24 & \textbf{53.52}
& \textbf{69.59} & 21.67 & 98.12 & 98.92 \\
\bottomrule
\end{tabular}%
}
\vspace{-0.3cm}
\end{table}

\begin{table}[t]
    \centering
    \vspace{-0.2cm}
    \caption{Analysis on number of 3D  tracks (left) and patch dimension in Track-the-Noise step (right).}
    \label{tab:analysis_points_patch}
    \begin{minipage}[t]{0.49\linewidth}
        \centering
        \resizebox{\linewidth}{!}{%
        \begin{tabular}{lcccc}
        \toprule
        Points & PSNR $\uparrow$ & SSIM $\uparrow$ & EPE $\downarrow$ & CLIP $\uparrow$ \\
        \midrule
        $N=1024$ & \textbf{19.28} & \textbf{0.78} & \textbf{2.26} & \textbf{0.9385} \\
        $N=512$  & 19.04 & 0.78 & 2.38 & 0.9373 \\
        $N=256$  & 18.44 & 0.77 & 2.78 & 0.9347 \\
        \bottomrule
        \end{tabular}%
        }
    \end{minipage}%
    \hfill
    \begin{minipage}[t]{0.49\linewidth}
        \centering
        \resizebox{\linewidth}{!}{%
        \begin{tabular}{lcccc}
        \toprule
        Patch & PSNR $\uparrow$ & SSIM $\uparrow$ & EPE $\downarrow$ & CLIP $\uparrow$ \\
        \midrule
        $R = 1.0$ & 16.95 & 0.71 & 5.34 & 0.9201 \\
        $R = 2.0$ & 19.28 & 0.78 & \textbf{2.26} & \textbf{0.9385} \\
        $R = 3.0$ & \textbf{19.40} & \textbf{0.79} & 2.28 & 0.9379 \\
        \bottomrule
        \end{tabular}%
        }
    \end{minipage}
\end{table}

\noindent\textbf{3D-sphere noise.} 
We validate 3D-sphere noise sampling in \Cref{fig:sphere} by ablating it and using random noise instead. Without sphere-induced cross-frame noise consistency, the model hallucinates implausible objects and inconsistent layouts, whereas enabling sphere sampling (Move-the-World) yields realistic, coherent motion. Quantitatively on MoveBench single-object (\Cref{tab:movebench_ablation}(a)), sphere noise improves PSNR (19.28 vs 19.10) and EPE (2.26 vs 2.42) and consistently boosts VBench-I2V metrics (consistency, aesthetics, reduced flicker, smoother motion), confirming it is a key component.

\noindent\textbf{Degradation rate.}
We observe that our 3D-grounded motion noise preserves spatial Gaussian characteristics but lacks temporal Gaussianity. 
To mitigate this issue, we blend the constructed motion-noise with pure Gaussian noise through alpha blending, using a randomly sampled degradation rate $\theta \in [0, 0.5]$.
In \Cref{tab:movebench_ablation}~(b) and \Cref{fig:degra_qual}, a degradation rate of $\theta = 0.2$ provides the best balance between visual quality  (PSNR, SSIM, Aesthetics Quality, Imaging Quality) and motion fidelity (EPE, Temporal Flickering, Motion Smoothness). We adopt $\theta = 0.2$ as the default in our experiments.


\noindent\textbf{Point tracks.}
As we use TAPIP3D~\cite{tapip3d} to extract 1024 points for each video during preprocessing, and then vary the total of point tracks during training, we provide an experiment to validate the robustness of our method when the number of point tracks are varied. 
\Cref{tab:analysis_points_patch} (left) and Fig.~\ref{fig:supp_numtrack_qual} provides the performance of our model on different number of 3D point tracks.
The results show that using $512$ and $1024$ point tracks yield similar PSNR, SSIM, and CLIP values, while using $1024$ leads to more favorable EPEs. Using $256$ points yields less competitive results, but the performance remains close to the $512$ and $1024$ case. 

\noindent\textbf{Patch size.}
We analyze the influence of the patch dimension ($2R - 1$) when propagating noise using point tracks as described in the Track-the-Noise step (\Cref{sec:tracknoise}). 
The results are in \Cref{tab:analysis_points_patch} (right) and Fig.~\ref{fig:supp_radius_qual}. 
It can be seen that using larger patch size ($R = 2$) yields the best EPE and CLIP while setting $R = 3$ yields the best PSNR and SSIM.  
Given the minor improvement in PSNR and SSIM, we opt for $R = 2$ as our default setting. 


\section{Conclusion}
\label{sec:conclusion}
We presented UniCaMo, an image-to-video generation framework that unifies camera motion and object motion through a 3D-grounded motion-consistent Gaussian noise representation.
By combining noise warping based on 3D point tracks and with 3D-sphere noise sampling, UniCaMo produces motion-consistent noise that guides video diffusion without modifying the underlying model architecture.
Our results on MoveBench suggest that structured noise initialization is an effective and efficient way for controllable video generation, enabling independent and joint control over motion.

\FloatBarrier


{
\small
\bibliographystyle{plainnat}
\bibliography{main}
}









\appendix

\clearpage
\section*{Technical appendices and supplementary material}

\input{sup}



\end{document}

%% file: definitions.tex
\def\mA{\mathcal{A}}
\def\mB{\mathcal{B}}
\def\mC{\mathcal{C}}
\def\mD{\mathcal{D}}
\def\mE{\mathcal{E}}
\def\mF{\mathcal{F}}
\def\mG{\mathcal{G}}
\def\mH{\mathcal{H}}
\def\mI{\mathcal{I}}
\def\mJ{\mathcal{J}}
\def\mK{\mathcal{K}}
\def\mL{\mathcal{L}}
\def\mM{\mathcal{M}}
\def\mN{\mathcal{N}}
\def\mO{\mathcal{O}}
\def\mP{\mathcal{P}}
\def\mQ{\mathcal{Q}}
\def\mR{\mathcal{R}}
\def\mS{\mathcal{S}}
\def\mT{\mathcal{T}}
\def\mU{\mathcal{U}}
\def\mV{\mathcal{V}}
\def\mW{\mathcal{W}}
\def\mX{\mathcal{X}}
\def\mY{\mathcal{Y}}
\def\mZ{\mathcal{Z}} 

\def\bbN{\mathbb{N}} 
\def\bbR{\mathbb{R}} 
\def\bbP{\mathbb{P}} 
\def\bbQ{\mathbb{Q}} 
\def\bbE{\mathbb{E}}

\def\1n{\mathbf{1}_n}
\def\0{\mathbf{0}}
\def\1{\mathbf{1}}

\def\A{{\bf A}}
\def\B{{\bf B}}
\def\C{{\bf C}}
\def\D{{\bf D}}
\def\E{{\bf E}}
\def\F{{\bf F}}
\def\G{{\bf G}}
\def\H{{\bf H}}
\def\I{{\bf I}}
\def\J{{\bf J}}
\def\K{{\bf K}}
\def\L{{\bf L}}
\def\M{{\bf M}}
\def\N{{\bf N}}
\def\O{{\bf O}}
\def\P{{\bf P}}
\def\Q{{\bf Q}}
\def\R{{\bf R}}
\def\S{{\bf S}}
\def\T{{\bf T}}
\def\U{{\bf U}}
\def\V{{\bf V}}
\def\W{{\bf W}}
\def\X{{\bf X}}
\def\Y{{\bf Y}}
\def\Z{{\bf Z}}

\def\a{{\bf a}}
\def\b{{\bf b}}
\def\c{{\bf c}}
\def\d{{\bf d}}
\def\e{{\bf e}}
\def\f{{\bf f}}
\def\g{{\bf g}}
\def\h{{\bf h}}
\def\i{{\bf i}}
\def\j{{\bf j}}
\def\k{{\bf k}}
\def\l{{\bf l}}
\def\m{{\bf m}}
\def\n{{\bf n}}
\def\o{{\bf o}}
\def\p{{\bf p}}
\def\q{{\bf q}}
\def\r{{\bf r}}
\def\s{{\bf s}}
\def\t{{\bf t}}
\def\u{{\bf u}}
\def\v{{\bf v}}
\def\w{{\bf w}}
\def\x{{\bf x}}
\def\y{{\bf y}}
\def\z{{\bf z}}

\def\balpha{\mbox{\boldmath{$\alpha$}}}
\def\bbeta{\mbox{\boldmath{$\beta$}}}
\def\bdelta{\mbox{\boldmath{$\delta$}}}
\def\bgamma{\mbox{\boldmath{$\gamma$}}}
\def\blambda{\mbox{\boldmath{$\lambda$}}}
\def\bsigma{\mbox{\boldmath{$\sigma$}}}
\def\btheta{\mbox{\boldmath{$\theta$}}}
\def\bomega{\mbox{\boldmath{$\omega$}}}
\def\bxi{\mbox{\boldmath{$\xi$}}}
\def\bnu{\mbox{\boldmath{$\nu$}}}                                  
\def\bphi{\mbox{\boldmath{$\phi$}}}
\def\bmu{\mbox{\boldmath{$\mu$}}}

\def\bDelta{\mbox{\boldmath{$\Delta$}}}
\def\bOmega{\mbox{\boldmath{$\Omega$}}}
\def\bPhi{\mbox{\boldmath{$\Phi$}}}
\def\bLambda{\mbox{\boldmath{$\Lambda$}}}
\def\bSigma{\mbox{\boldmath{$\Sigma$}}}
\def\bGamma{\mbox{\boldmath{$\Gamma$}}}
                                  
\newcommand{\myprob}[1]{\mathop{\mathbb{P}}_{#1}}

\newcommand{\myexp}[1]{\mathop{\mathbb{E}}_{#1}}

\newcommand{\mydelta}[1]{1_{#1}}

\newcommand{\myminimum}[1]{\mathop{\textrm{minimum}}_{#1}}
\newcommand{\mymaximum}[1]{\mathop{\textrm{maximum}}_{#1}}    
\newcommand{\mymin}[1]{\mathop{\textrm{minimize}}_{#1}}
\newcommand{\mymax}[1]{\mathop{\textrm{maximize}}_{#1}}
\newcommand{\mymins}[1]{\mathop{\textrm{min.}}_{#1}}
\newcommand{\mymaxs}[1]{\mathop{\textrm{max.}}_{#1}}  
\newcommand{\myargmin}[1]{\mathop{\textrm{argmin}}_{#1}} 
\newcommand{\myargmax}[1]{\mathop{\textrm{argmax}}_{#1}} 
\newcommand{\myst}{\textrm{s.t. }}

\newcommand{\denselist}{\itemsep -1pt}
\newcommand{\sparselist}{\itemsep 1pt}

\definecolor{pink}{rgb}{0.9,0.5,0.5}
\definecolor{purple}{rgb}{0.5, 0.4, 0.8}   
\definecolor{gray}{rgb}{0.3, 0.3, 0.3}
\definecolor{mygreen}{rgb}{0.2, 0.6, 0.2}

\newcommand{\cyan}[1]{\textcolor{cyan}{#1}}
\newcommand{\red}[1]{\textcolor{red}{#1}}  
\newcommand{\blue}[1]{\textcolor{blue}{#1}}
\newcommand{\magenta}[1]{\textcolor{magenta}{#1}}
\newcommand{\pink}[1]{\textcolor{pink}{#1}}
\newcommand{\green}[1]{\textcolor{green}{#1}} 
\newcommand{\gray}[1]{\textcolor{gray}{#1}}    
\newcommand{\mygreen}[1]{\textcolor{mygreen}{#1}}    
\newcommand{\purple}[1]{\textcolor{purple}{#1}}       

\definecolor{greena}{rgb}{0.4, 0.5, 0.1}
\newcommand{\greena}[1]{\textcolor{greena}{#1}}

\definecolor{bluea}{rgb}{0, 0.4, 0.6}
\newcommand{\bluea}[1]{\textcolor{bluea}{#1}}
\definecolor{reda}{rgb}{0.6, 0.2, 0.1}
\newcommand{\reda}[1]{\textcolor{reda}{#1}}

\def\changemargin#1#2{\list{}{\rightmargin#2\leftmargin#1}\item[]}
\let\endchangemargin=\endlist
                                               
\newcommand{\cm}[1]{}

\newcommand{\mhoai}[1]{{\color{magenta}\textbf{[Hoai: #1]}}}

\newcommand{\mtodo}[1]{{\color{red}$\blacksquare$\textbf{[TODO: #1]}}}
\newcommand{\myheading}[1]{\vspace{0.5ex}\noindent \textbf{#1}}
\newcommand{\htimesw}[2]{\mbox{$#1$$\times$$#2$}}


%
%
%

\newcommand{\Sref}[1]{Sec.~\ref{#1}}
\newcommand{\Eref}[1]{Eq.~(\ref{#1})}
\newcommand{\Fref}[1]{Fig.~\ref{#1}}
\newcommand{\Tref}[1]{Table~\ref{#1}}

%% file: sup.tex

\section{Implementation Details}
\label{sec:supp_impl}

\subsection{Inference Pipeline}
\label{sec:infer_scheme}

\begin{figure}[!ht]
    \centering
    \includegraphics[width=0.9\textwidth]{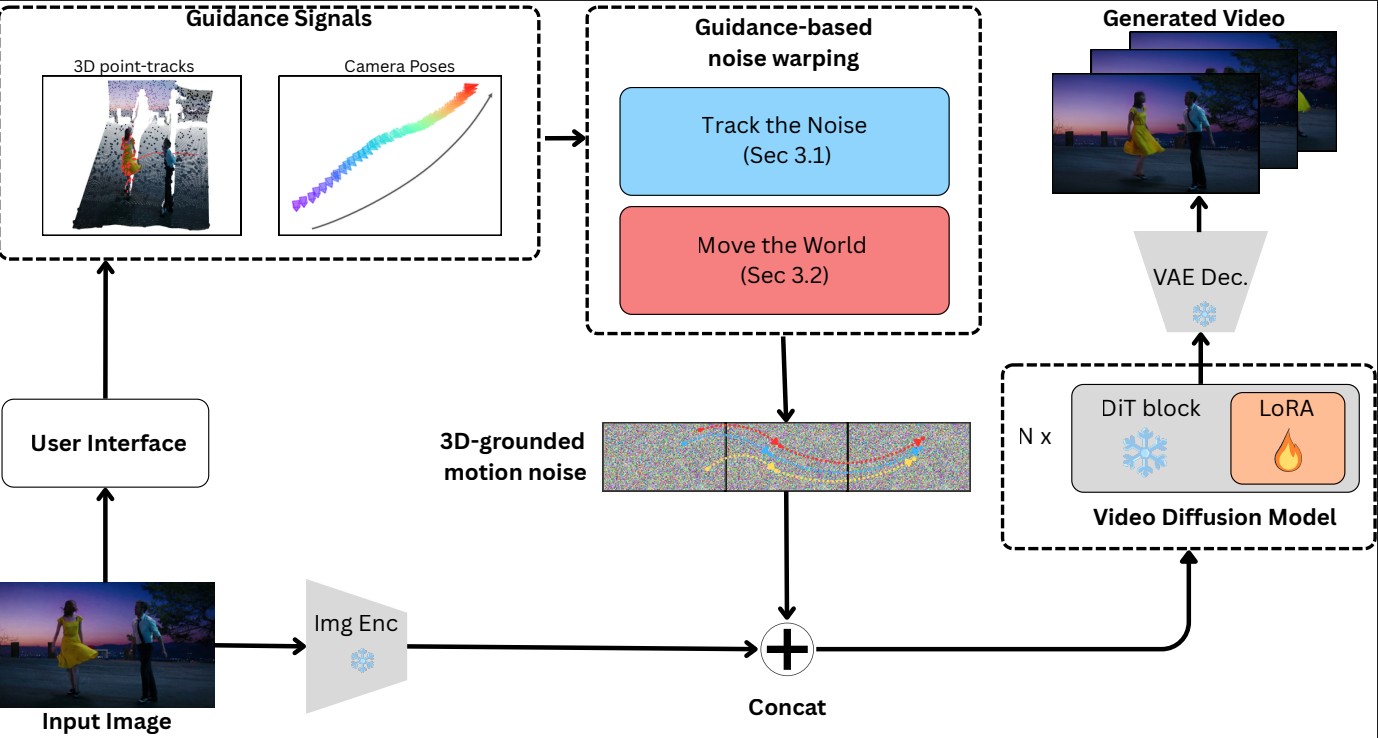}
    \caption{Overview of our inference pipeline.}
    \label{fig:inference_scheme}
\end{figure}

Fig.~\ref{fig:inference_scheme} illustrates our inference pipeline.
Given a single input image, we first reconstruct the scene using ViPE~\cite{huang2025vipe} to obtain a 3D point cloud, depth map, and camera intrinsics. The user then specifies desired object and camera motions through our interactive GUI (Sec.~\ref{sec:UI}), which exports guidance signals in the form of 3D point tracks and camera poses. These signals are fed into our Guidance-based Noise Warping module, consisting of Track-the-Noise (Sec.~\ref{sec:tracknoise}) for encoding object motion and Move-the-World (Sec.~\ref{sec:sphere}) for encoding camera motion, to construct 3D-grounded motion noise. The constructed noise is then concatenated with the encoded image latent and passed to a LoRA-adapted video diffusion model. Finally, the output latent is decoded via a VAE decoder to produce the generated video with faithful object and camera motion control.

\subsection{Implementation Details of Track-the-Noise}
\label{sec:track_the_noise_imple}
\paragraph{Patch copy--paste along trajectories.}
We initialize the first-frame latent noise as i.i.d. Gaussian,
\begin{align}
\mathbf{Z}_T^{1} \sim \mathcal{N}(\mathbf{0}, \mathbf{I}).
\end{align}
For each tracked point $i$ and frame $t$, let $(u_i^t, v_i^t)$ be its (integer) latent coordinates.
To propagate an identifiable noise footprint along the trajectory, we copy a small square patch around the track location in frame $1$ to the corresponding location in frame $t$.
Using a patch radius $r$, we define a patch of  $(2r{-}1)\times(2r{-}1)$ pixels where the copy follows:
\begin{align}
\mathbf{Z}_T^{t}\!\left[u_i^t+\Delta u,\; v_i^t+\Delta v\right]
:=
\mathbf{Z}_T^{1}\!\left[u_i^1+\Delta u,\; v_i^1+\Delta v\right].
\end{align}
where $\Delta u$ and $\Delta v$ are the pixel offsets. 
For $r=2$, the patch size is $3\times 3$, and the offset is
in $\{-1,0,1\}$. 
By transporting the same local noise pattern across time, motion becomes \emph{visible} to the latents so that visual features in the diffusion process are encouraged to follow the specified trajectory.
When multiple trajectories project to the same latent location at time $t$, naive composition yields ambiguous assignments and can introduce inconsistent motion cues. We then utilize Depth-aware Collision Tracking policies that mentioned in \ref{sec:tracknoise} to resolve overlap.

\subsection{Implementation Details of Move-the-World}
\label{sec:move_the_world_imple}
\paragraph{Ray construction (pixel $\rightarrow$ world ray).}
For each pixel $(p_x,p_y)$ (we use pixel centers), we form a unit ray direction in camera coordinates using intrinsics $(f_x,f_y,c_x,c_y)$:
\begin{align}
\mathbf{d}_{c}(p_x,p_y) 
&= 
\frac{1}{\left\| \left(\frac{p_x-c_x}{f_x},\,\frac{p_y-c_y}{f_y},\,1\right) \right\|_2}
\begin{bmatrix}
\frac{p_x-c_x}{f_x}\\[2pt]
\frac{p_y-c_y}{f_y}\\[2pt]
1
\end{bmatrix},
\end{align}
and then apply a camera-to-world transformation to determine the ray direction in the world space:
\begin{align}
\mathbf{d}_{w}(p_x,p_y) 
&= \mathbf{R}^t \, \mathbf{d}_{c}(p_x,p_y),
\end{align}
where $\mathbf{R}^t$ is the rotation from $\mathbf{T}_{c\rightarrow w}^t$.
The origin of the ray is at the camera location. 

\paragraph{Ray-sphere intersection and spherical UV.}
Let $\mathbf{S}$ be the sphere center with radius $r$. 
Let $\mathbf{o}^t$ be the ray origin and $\mathbf{d}_w$ be the ray direction in the world space. We define the direction from the sphere center $\mathbf{S}$ to the ray origin as $\mathbf{O}^t=\mathbf{o}^t-\mathbf{S}$.
We perform ray-sphere intersection by solving a quadratic equation.
With $b=\langle \mathbf{d}_w,\mathbf{O}^t\rangle$ and $c=\|\mathbf{O}^t\|_2^2-r^2$, the (non-negative) discriminant is
\begin{equation}
\Delta = b^2 - c,
\end{equation}
and we take the exiting intersection (consistent with the camera being inside or near the sphere):
\begin{equation}
\tau = -b + \sqrt{\Delta} .
\end{equation}
The hit point and outward normal are
\begin{equation}
\mathbf{P} = \mathbf{o}^t + \tau \mathbf{d}_w,
\qquad \mathrm{and} \qquad
\mathbf{n} = \frac{\mathbf{P}-\mathbf{S}}{r}.
\end{equation}
We convert the normal to spherical coordinates using longitude--latitude mapping:
\begin{align}
u &= \mathrm{fract}\!\left(\frac{\mathrm{atan2}(\mathbf{n}_x, \mathbf{n}_z)}{2\pi} + \frac{1}{2}\right),
\\
v &= \frac{\arccos(\mathrm{clip}(\mathbf{n}_y,-1,1))}{\pi},
\end{align}
where $\mathrm{fract}(\cdot)$ wraps $u$ into $[0,1)$.

\paragraph{Sampling sphere noise and downsampling to latent.}
We sample the Gaussian noise map at the computed spherical coordinates using nearest-neighbor lookup,
yielding a per-frame noise image.
This yields a 3D-consistent noise field while preserving the spatial Gaussianity per frame.

\section{Additional Qualitative Visualizations}
\label{sec:supp_qual}

\subsection{Additional Qualitative Comparisons on MoveBench Dataset}
\label{sec:supp_morequal}

\begin{figure}[!htbp]
    \centering
    \includegraphics[width=\linewidth]{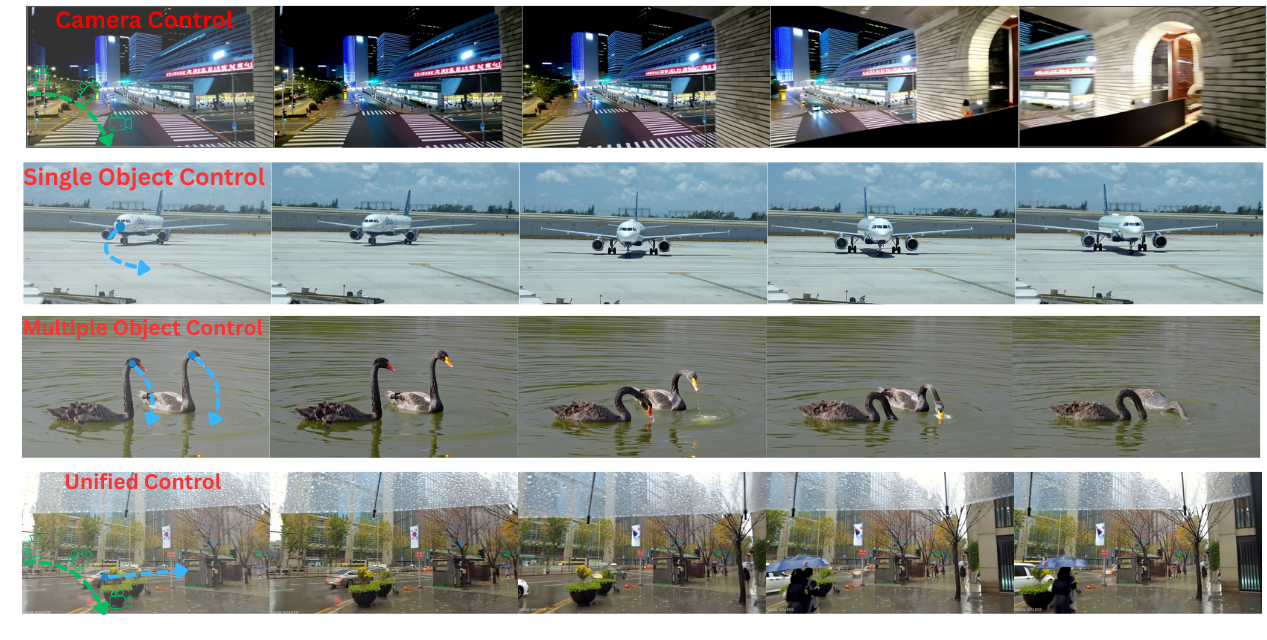}
    \caption{We demonstrate control of the camera motion, single-object movement, multiple-object movement, and unified control of camera and object motion in diverse scenes.}
    \label{fig:old_teaser}
\end{figure}

\begin{figure}[!ht]
    \centering
    \includegraphics[width=\linewidth]{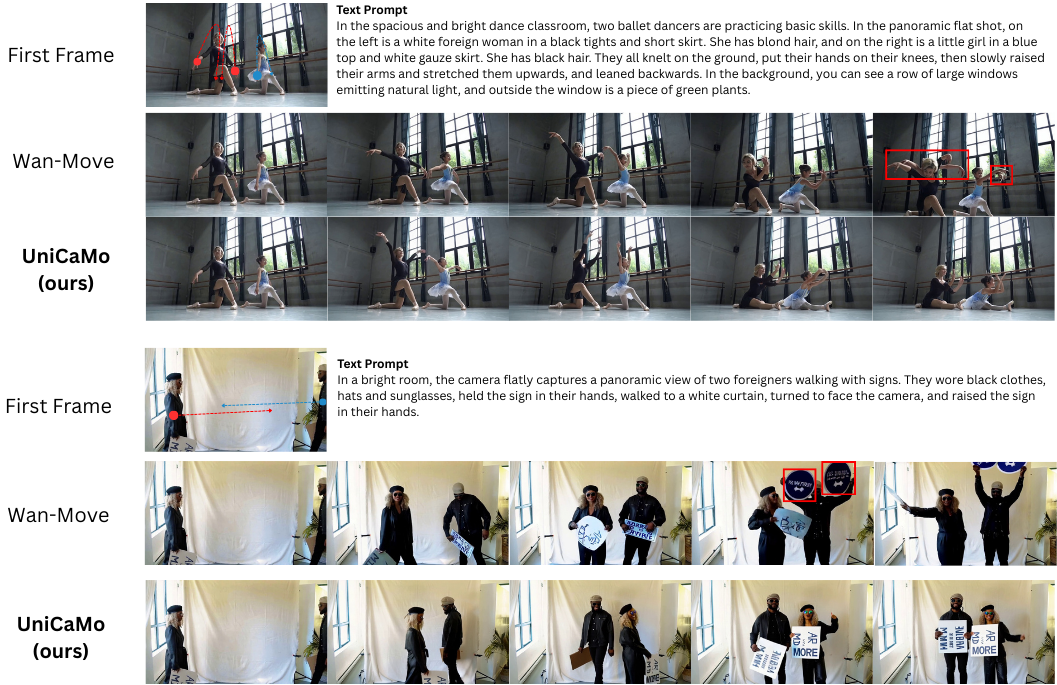}
    \caption{Additional qualitative comparison on MoveBench (1/3). Our method produces more physically plausible motion and better preserves scene structure compared to Wan-Move~\cite{chu2025wan}.}
    \label{fig:supp_qual1}
\end{figure}

\begin{figure}[!ht]
    \centering
    \includegraphics[width=\linewidth]{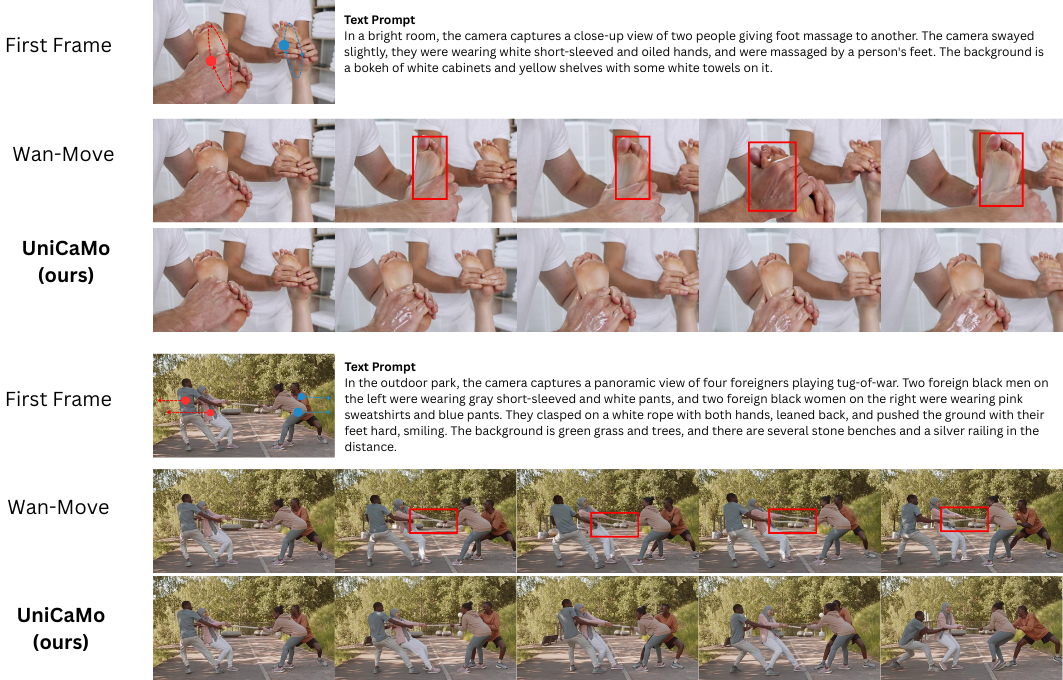}
    \caption{Additional qualitative comparison on MoveBench (2/3). Under complex object interactions, UniCaMo better follows the target trajectories and demonstrates stronger prompt adherence while maintaining visual quality.}
    \label{fig:supp_qual2}
\end{figure}

\begin{figure}[!ht]
    \centering
    \includegraphics[width=\linewidth]{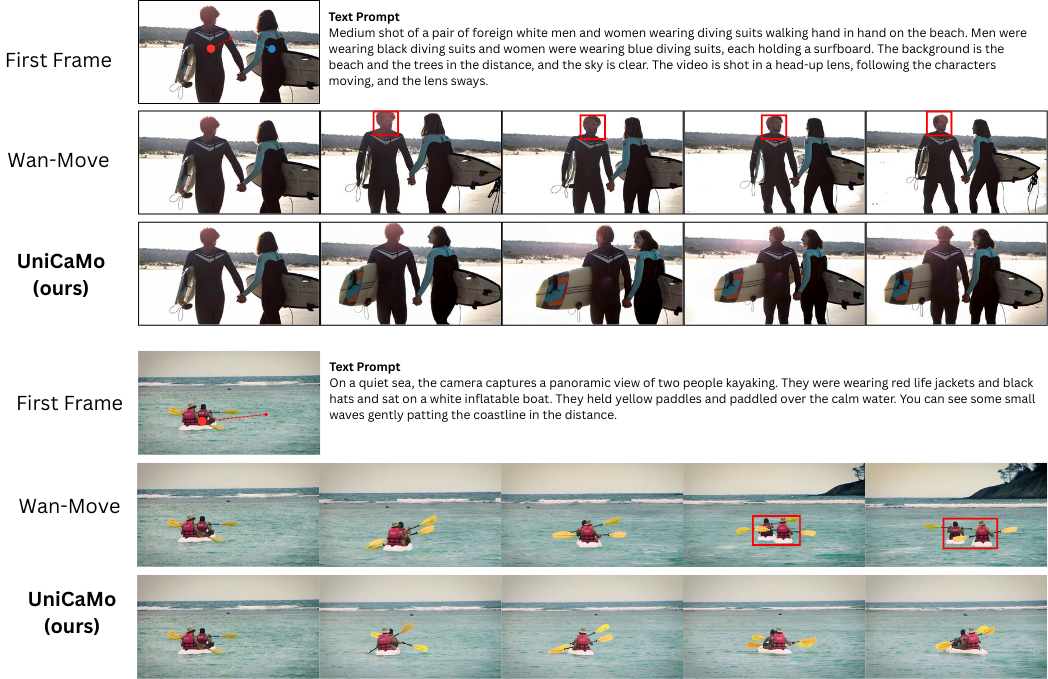}
    \caption{Additional qualitative comparison on MoveBench (3/3).}
    \label{fig:supp_qual3}
\end{figure}

We provide additional qualitative comparisons between UniCaMo and Wan-Move~\cite{chu2025wan} on different control tasks (see~\cref{fig:old_teaser}), and on  challenging samples from MoveBench in Fig.~\ref{fig:supp_qual1}, Fig.~\ref{fig:supp_qual2}, and Fig.~\ref{fig:supp_qual3}. These examples feature complex scenes with multiple interacting objects. Under such challenging conditions, Wan-Move often results with incorrect motion directions, implausible object deformations, or failure to follow the text prompt faithfully. In contrast, UniCaMo consistently generates videos with more physically plausible motion dynamics and better prompt adherence, demonstrating the effectiveness of our 3D-grounded motion-consistent noise in encoding both object and camera motion cues. 

\subsection{Additional Qualitative for Ablation Studies}
\label{sec:supp_table2}

\begin{figure}[!tbp]
    \centering
    \includegraphics[width=0.8\linewidth]{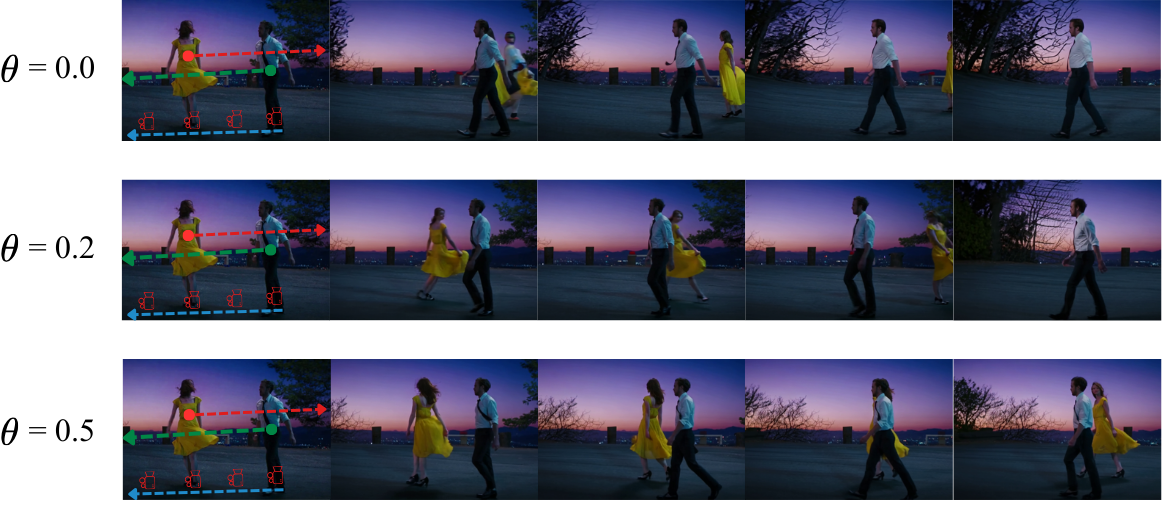}
    \caption{Effect of degradation rate on generated samples. A lower degradation rate ($\theta = 0.0$) offers more precise trajectory control but introduces visible artifacts. Conversely, a higher degradation rate ($\theta = 0.5$) yields visually plausible videos but reduces 3D trajectory accuracy. We adopt $\theta = 0.2$ as the default, which achieves the best trade-off between motion fidelity and visual quality.}
    \label{fig:degra_qual}
\end{figure}

\begin{figure}[t]
    \centering
    \includegraphics[width=0.9\linewidth]{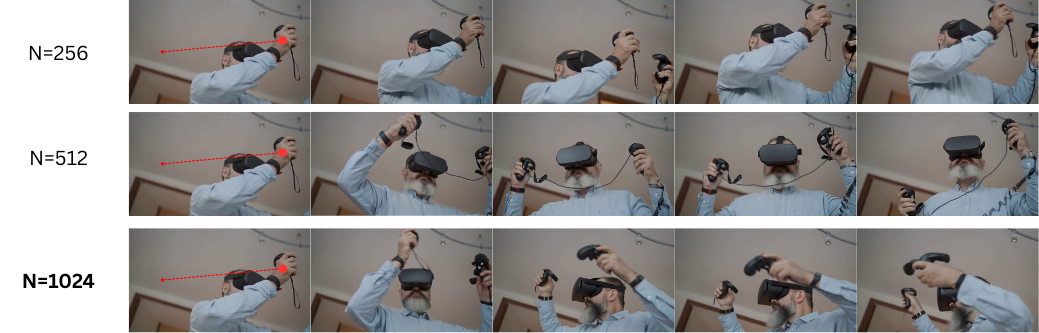}
    \caption{Qualitative results of our proposed UniCaMo method using different number of 3D point tracks.}
    \label{fig:supp_numtrack_qual}
\end{figure}

\begin{figure}[t]
    \centering
    \includegraphics[width=0.9\linewidth]{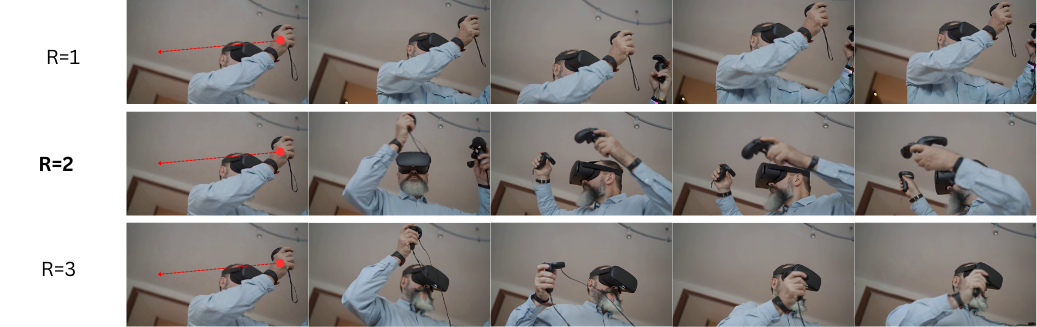}
    \caption{Qualitative results of our proposed UniCaMo method using different configuration of path dimension $R$.}
    \label{fig:supp_radius_qual}
\end{figure}

In this section, we provide more qualitative results on the performance of our UniCaMo model on the degradation rate (see~\cref{fig:degra_qual}), on different number of 3D point tracks (see Fig.~\ref{fig:supp_numtrack_qual}) and
on different patch dimensions (see Fig.~\ref{fig:supp_radius_qual}).

\clearpage
\subsection{User Interface}
\label{sec:UI}
\begin{figure}[h!]
    \centering
    \includegraphics[width=\linewidth]{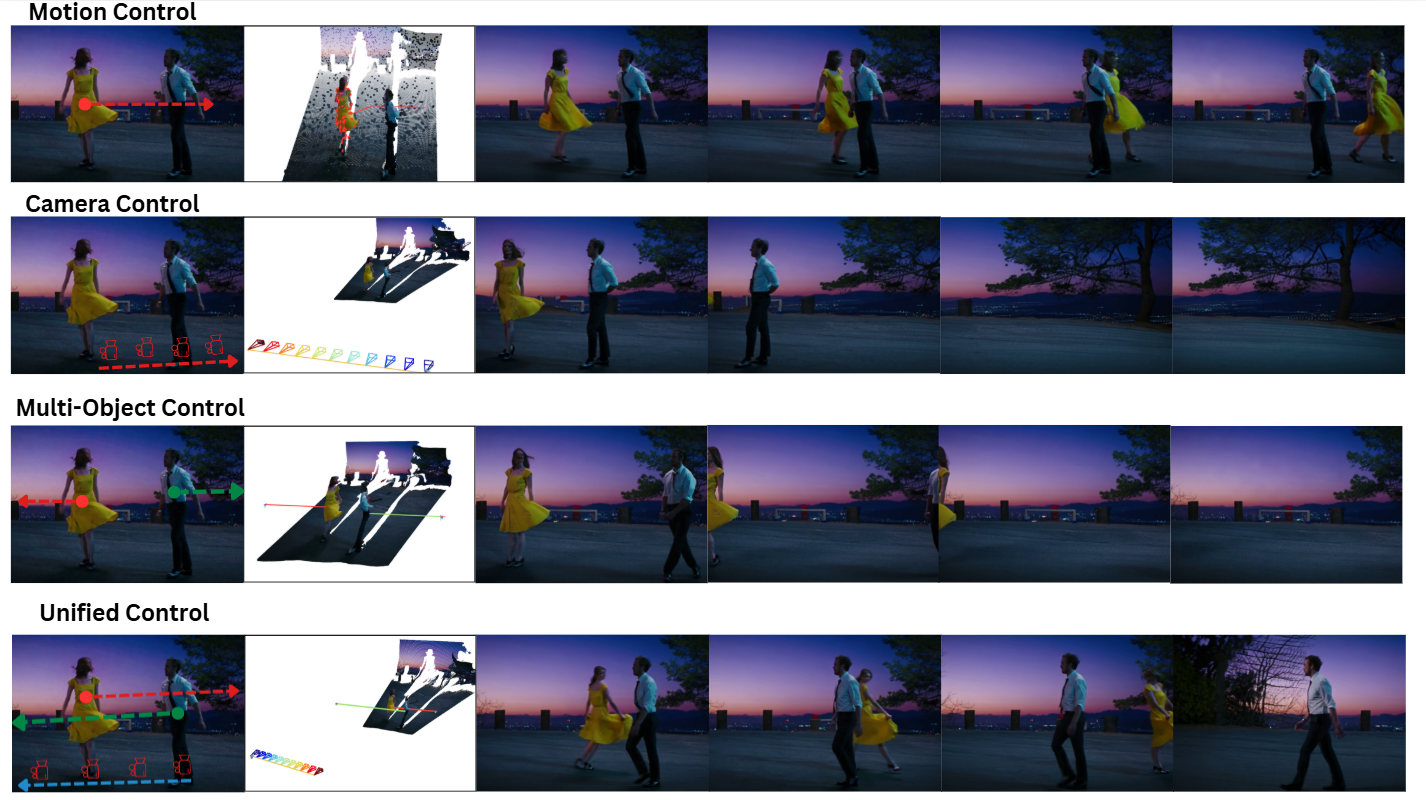}
    \caption{Visualization of our interactive GUI for unified camera and object motion control. Users select objects via polygon masks on the 2D image and edit 3D object trajectories and camera paths using interactive gizmos in the 3D view. Notice how 3D point-tracks allow us to define the exact 3D trajectory of the woman in yellow dress (go behind the man).}
    \label{fig:ui_demo}
\end{figure}

To support unified control of both object and camera motion at inference time, we implement an interactive graphical user interface (GUI) that enables users to define 3D object trajectories and 3D camera trajectories through an intuitive two-phase workflow. Fig.~\ref{fig:ui_demo} shows a visualization of our GUI and results.

\paragraph{Data Processing.}
Given a single input image, we first run ViPE~\cite{huang2025vipe} to estimate monocular depth and camera intrinsics. Using the estimated depth map and camera parameters, we reconstruct a dense 3D point cloud of the scene. This point cloud serves as the spatial context for all subsequent editing operations.

\paragraph{2D Object Selection.}
In the first phase, the GUI displays the input image as a 2D canvas. The user selects the object(s) of interest by drawing polygon masks directly on the image. Multiple objects can be selected as separate groups, each receiving its own mask. From each object mask, we proportionally sample 3D point tracks: for an object whose mask covers $n$ pixels out of the total $H \times W$ image, we allocate $\max(1, \lfloor n / (H \times W) \times N \rfloor)$ of the $N = 1024$ total point tracks. The remaining tracks are sampled uniformly from regions outside all masks to represent background motion. Each sampled pixel is mapped to its corresponding 3D world-space coordinate via the reconstructed point cloud.

\paragraph{3D Object Motion Editing.}
In the second phase, the GUI transitions to a 3D view showing the reconstructed point cloud alongside the sampled point tracks. The user defines target object trajectories by placing keyframes along a desired 3D path using interactive translation and rotation gizmos (6-DoF transform controls). Intermediate frames are obtained via linear interpolation between keyframes, producing continuous target motion trajectories for all points within each object group.

\paragraph{Camera Trajectory Editing.}
Camera motion is edited through a separate set of 6-DoF gizmos that control the virtual camera's position and orientation at user-specified keyframes. The camera trajectory is visualized as a sequence of frustums connected by a spline curve, allowing the user to intuitively design complex camera paths including panning, tilting, dollying, and orbiting. As with object motion, intermediate camera poses are linearly interpolated between keyframes.

\paragraph{Export and Video Generation.}
Once the user finalizes both object and camera trajectories, the GUI exports a single \texttt{.npz} file containing: (i) 3D point track coordinates of shape $(T, N, 3)$, (ii) per-frame visibility flags of shape $(T, N)$, (iii) camera intrinsics of shape $(T, 3, 3)$, and (iv) camera extrinsics (world-to-camera) of shape $(T, 4, 4)$. These exported signals are directly fed into our Guidance-based Noise Warping module (Sec.~\ref{sec:tracknoise} and Sec.~\ref{sec:sphere}) to construct the 3D-grounded motion-consistent noise, which then initializes the video diffusion model for generation.

\section{Discussion, Limitations, and Future Work}
While our 3D-grounded, motion-consistent noise enables flexible control of object and camera motion, the method has limitations. First, the Track-the-Noise stage depends heavily on accurate 3D point tracks and camera poses, so tracking or calibration errors can degrade controllability. Second, the Move-the-World stage uses a coarse 3D sphere proxy, which may fail to capture complex background geometry. Future improvements in 3D tracking, depth estimation, and noise representations would further enhance our approach.